# HuTO: une Ontologie Temporelle Narrative pour les Applications du Web Sémantique

Papa Fary Diallo[1,2,3], Olivier Corby[1,2], Isabelle Mirbel[2]

Moussa Lo[3] and Seydina M. Ndiaye[3]

[1] INRIA Sophia Antipolis, FRANCE,
`{papa-fary.diallo, olivier.corby}@inria.fr`
[2] Univ. Nice Sophia Antipolis, CNRS, I3S, UMR 7271, FRANCE,
`isabelle.mirbel@unice.fr`
[3] Université Gaston Berger - UFR SAT - LANI, SENEGAL,
`{moussa.lo, seydina.ndiaye}@ugb.edu.sn`

**Abstract** : Un défi majeur en informatique est la modélisation et le raisonnement sur les données temporelles. Ce travail est devenu encore plus important avec l'émergence du Web sémantique où de grandes quantités données hétérogènes sont manipulées. Ces données comportent souvent des informations temporelles informelles, semi-formelles ou formelles qui doivent être interprétées par les agents logiciels. Dans cet article nous présentons notre ontologie, Humain Time Ontologie (HuTO), une ontologie en RDFS pour annoter des ressources en RDF et représenter les expressions narratives temporelles. Une des contributions majeures de HuTO est la modélisation des intervalles non-convexes c'est-à-dire les intervalles répétitifs comme `tous les mercredi` mais également la possibilité d'écrire des requêtes sur ce type d'intervalle. HuTO intègre aussi des règles de normalisation et de raisonnement pour expliciter certaines informations temporelles. HuTO propose aussi une approche qui permet de garder distincte la dimension temporelle et les annotations du domaine métier. Cela facilite la recherche d'informations qu'elles soient temporelles ou non.

**Mots-clés** :Ontologies Temporelles, Web Sémantique, RDFS, SPARQL, Règles.

## 1 Introduction

Les phénomènes temporels ont de nombreuses facettes qui sont étudiées par différentes communautés. Ainsi, la dimension temporelle des données est aussi étudiée dans le domaine de l'informatique où il y a un besoin croissant de modéliser des systèmes calendaires, des événements répétitifs et des faits qui sont vrais pour un certain temps et faux par ailleurs. C'est le cas des Systèmes d'Information qui doivent faire face au problème des données obsolètes. En Intelligence Artificielle, des modèles abstraits ont été proposés pour pouvoir raisonner sur des concepts temporels. Dans ce domaine, Allen (Allen, 1984, 1981) a présenté un modèle de calcul entre les intervalles de temps qui a influencé les travaux sur la modélisation du temps. Ces travaux de Allen ont été étendus aux intervalles non-convexes (intervalles répétitifs) par Ladkin (Ladkin, 1987). Dans le Traitement Automatique des Langages Naturels (TALN) les modèles développés cherchent à extraire les expressions temporelles mais aussi leur sémantique en langue naturelle. Ainsi, un défi important dans le domaine de l'informatique est la représentation et le raisonnement sur des informations temporelles. L'intérêt de ce travail est de plus en plus important maintenant avec l'émergence du Web sémantique où de gros volumes de données hétérogènes sont manipulés.

Dans le domaine du Web sémantique sont présentes à la fois des notions temporelles informelles, semi-formelles et formelles qui doivent être comprises par les agents logiciels. Nous distinguons deux axes de travail: la modélisation d'expression temporelle et l'annotation temporelle des données. La modélisation d'expression temporelle permet de modéliser une date, un intervalle, des notions temporelles répétitives, relatives ou absolues, etc. L'annotation temporelle des données permet la représentation de notions temporelles de façon à annoter des connaissances (exprimées sous forme de triplet en RDF) et cela en conservant l'évolution des données (changement de valeur) dans le temps. Pour cela, le Web sémantique repose sur des ontologies qui sont une *spécification explicite et formelle d'une conceptualisation partagée* (Studer et al., 1998). Ainsi, l'objectif principal de ce travail est de proposer une ontologie pour représenter des notions temporelles et annoter temporellement des données.

Dans (Diallo et al., 2011, 2014) nous avons développé une ontologie socioculturelle et une plateforme de partage et de co-construction de connaissances sur les communautés sénégalaises. La manipulation de ces données socioculturelles fait intervenir beaucoup de notions temporelles. Ainsi dans cet article, nous présentons notre ontologie temporelle, Human Time Ontology (HuTO), et nous illustrons son utilisation sur des données extraites de cette plateforme.

Ce document continue par un état de l'art dans lequel nous présenterons les travaux sur la modélisation des notions temporelles et l'annotation temporelle des données dans le Web sémantique. Ensuite la troisième partie détaillera notre proposition d'ontologie: HuTO. En premier lieu nous présenterons les concepts de l'ontologie qui servent à modéliser la représentation d'énoncé de temps complexe. En deuxième lieu nous présenterons notre approche pour l'annotation temporelle des données. Dans la partie quatre, nous présenterons les raisonnements et les règles proposés dans HuTO. Dans la cinquième partie, nous montrerons des exemples de requêtes en SPARQL sur des connaissances temporellement annotées à l'aide de HuTO. Nous terminerons par une conclusion et des perspectives pour ce travail.

## 2   État de l'art

Plusieurs spécifications ont été proposées pour modéliser des expressions (énoncés) temporelles parmi lesquelles nous pouvons citer TimeML (Sauri, 2006), OWL-Time (Pan et Hobbs, 2005) (Pan, 2007) et CNTRO (Tao et al., 2010, 2011). TimeML est un langage d'annotation pour les informations temporelles dans des documents textuels utilisé dans le TALN. Il est basé sur un système de balises XML standard. L'inconvénient principal de ce langage est qu'il annote les événements et les expressions temporelles dans des segments textuels isolés, ce qui rend la recherche d'information plus difficile. TimeML ne permet pas non plus d'exprimer des expressions comme `every 3rd Monday`. OWL-Time est une ontologie temporelle qui permet de fournir une description temporelle de documents du Web et de Web services. CNTRO est une ontologie en OWL pour la modélisation des informations temporelles dans les récits et rapports cliniques. Ces deux ontologies permettent entre autres la modélisation d'intervalles non-convexes et la représentation des relations comme celles définies par Allen (Allen, 1984, 1981). Cependant, Pour modéliser des intervalles non-convexes CNTRO modélise la périodicité dans des chaînes de caractères d'où la perte de la sémantique. OWL-Time ne permet pas de modéliser les expressions humaines du temps comme le temps déictique.

Dans les langages du Web sémantique comme RDF, un énoncé (statement) est une relation binaire qui est utilisée pour relier deux individus (instances) ou un individu et une valeur. Or, pour introduire une dimension temporelle, il devient nécessaire de manipuler des relations ternaires. La modélisation des relations ternaires est un cas particulier d'une problématique plus générale qui est la modélisation et l'interrogation des relations n-aires dans le Web sémantique. Ainsi dans la littérature il existe des approches générales qui essaient de répondre à cette problématique comme l'approche des N-ary relations (W3C Working Group, 2006) qui propose l'introduction d'un blank node entre l'objet et le sujet du triplet. Ainsi, le blank node peut par exemple être temporellement annoté. Il existe aussi l'approche des graphes nommés qui permettent de contextualiser un ensemble de triplets en les regroupant dans un même graphe (URI) qui peut par exemple être temporellement annoté. Une autre approche est la Réification en RDF qui permet grâce à `rdf:Statement` d'ajouter d'autres informations sur un triplet comme des informations temporelles.

Il existe aussi des approches spécifiques à la modélisation temporelle comme 4D-Fluents (Welty et Fikes, 2006), une ontologie en OWL qui propose une approche basée sur les occurrents et les perdurants pour modéliser l'évolution temporelle des données. Dans cette approche les auteurs considèrent que tout objet a une partie temporelle et que ce sont ces parties temporelles qui sont en interaction. Il y a aussi l'approche de SOWL (Batsakis et Petrakis, 2011) qui étend le 4D-Fluents en y ajoutant les relations d'Allen. Il existe aussi l'approche Temporal RDF (Gutierrez et al., 2005) qui étend la Réification en RDF en ajoutant une dimension temporelle sur les données. Ainsi le graphe peut être accédé selon deux vues, selon qu'on s'intéresse à la temporalité ou aux connaissances du domaine modélisé. Dans (Rula et al., 2014), les auteurs proposent une approche générique pour extraire des informations temporelles du Web et de leur durée de validité. (Scheuermann et al., 2013) propose une approche empirique centrée sur les perspectives des utilisateurs ce qui a permis de définir différents modèles temporels.

Excepté Temporal RDF, le principal inconvénient pour les autres approches est la perte des relations directes entre les ressources pour l'ajout de l'information temporelle. Ainsi pour ajouter des informations temporelles les triplets sont cassés on ajoutant des ressources intermédiares (N-ary relations, la réification en RDF) ou déplacés vers des `timeSlices` (4D-Fluents).

## 3   HuTO

HuTO[1] est une ontologie formalisée en RDFS permettant l'annotation temporelle de ressources en RDF à l'aide d'expressions temporelles du langage courant. Cette ontologie permet également de définir des ancrages temporels liés au contexte et de capturer les changements temporels associés aux ressources annotées. Elle rend possible l'interrogation temporelle de la base de connaissances à l'aide de requêtes SPARQL. Plus précisément, HuTO permet de:

- Modéliser des expressions temporelles:
    1. **Explicites**: elles sont immédiatement ancrées; par exemple: 30 Août 2014, été 2014;
    2. **Déictiques**: elles forment une relation spécifique avec le temps du discours; par exemple: aujourd'hui, demain;
    3. **De durée**: elles indiquent un intervalle de temps; par exemple, 2 heures, 20 minutes;
    4. **Cycliques**: elles permettent de modéliser des dates répétitives; par exemple: chaque lundi, tous les deux mois;
    5. **Mixtes**: elles combinent les expressions pré-cités; par exemple deux mois l'année dernière.
- Normaliser les expressions temporelles afin de pouvoir leur appliquer des raisonnements et de pouvoir les interroger.

### 3.1   Date, Temps calendaire et granularité

Dans HuTO, les concepts principaux pour la datation sont `Datation` et `TemporalUnit` (Fig. 1). `Datation` est un concept abstrait (qui n'a pas d'instances directes) dont dérivent les concepts `Date` et `Duration`. Le concept `Date` permet de modéliser des dates comme celles du type `xsd:dateTime` excepté la partie fuseau horaire (exemples 1a et 1b). Le concept `Duration` permet de définir des durées comme celles du type `xsd:duration` (exemple 1d). Les granularités définies dans `TemporalUnit` vont de `Century` à `Second`.

---
[1] http://ns.inria.fr/huto/

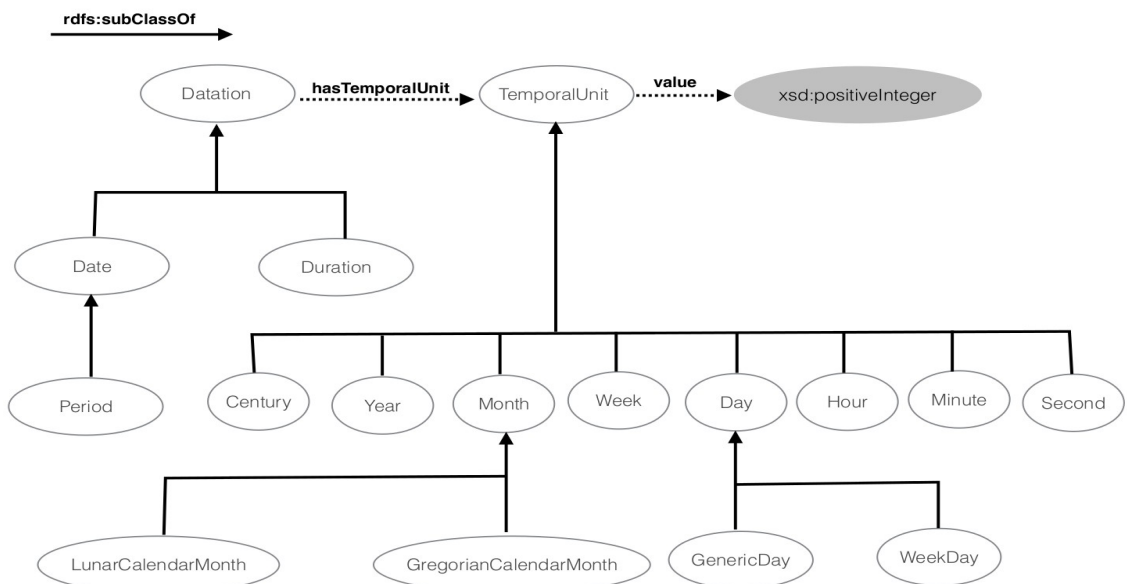

FIGURE 1 – Modélisation des types `Datation` et `TemporalUnit`.

Notons aussi que le concept `WeekDay` rassemble les jours de la semaine comme sous-concepts. Le concept `GenericDay` rassemble des sous-concepts comme `Today`, `Yesterday`, etc. Notons également la relation `hasContext` (exemple 1c), qui est utilisée pour contextualiser le concept `GenericDay`.

a. Date(Mardi 17 Février 2015 à 10H)
```
[a :Date;
   :hasHour  [a :Hour;
              :hour  10];
   :hasDay [a :Tuesday;
            :day 17];
   :hasMonth [a :February];
   :hasYear [a :Year;
             :year  2015]].
```

b. Date(15 04) "15 Avril"
```
[a :Date;
   :hasDay [a :Day;
            :day 15];
   :hasMonth
    [a :Month;
       :month 4]].
```

c. Date(Aujourd'hui)- Vendredi 29 Août 2014
```
[a :Date;
   :hasDay [a :Today;
            :hasContext
             [a :Date
                :hasDay
                 [a :Friday;
                    :day 29];
                :hasMonth
                 [a :August];
                :hasYear
                 [a :Year;
                    :year
                      2014]]]].
```

d. 2 heures 30 minutes
```
[a :Duration;
   :hasHour [a :Hour
             :value 2];
   :hasMinute [a :minute;
               :value 30]].
```

EXMPLE 1 – *Modélisation de notions de dates simples.*

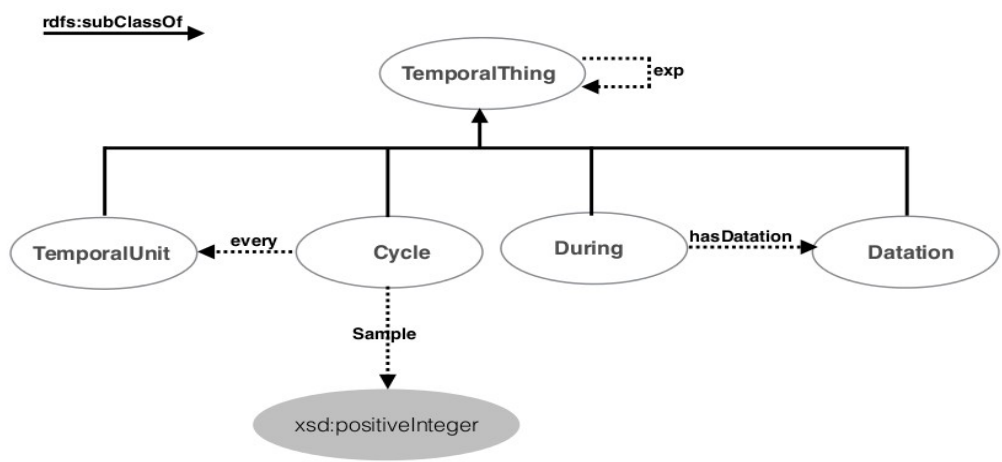

FIGURE 2 – Les concepts temporels de HuTO.

Notons que la propriété `hasTemporalUnit` (Fig. 1) est spécialisée par les propriétés `hasCentury`, `hasYear`, etc et la propriété `value` par `century`, `year` etc.

### 3.2 Instant, Intervalle et Durée

Un élément temporel peut être considéré comme un instant, un intervalle ou une durée. Nous avons fait le choix de représenter tous les éléments temporels comme des intervalles modélisés à l'aide du concept `During` (Fig. 2). De ce fait, si pour un intervalle, la date de fin ou la durée n'est pas spécifiée alors l'intervalle considéré est celui de l'unité de la date. Par exemple, la date `Vendredi 15 Août 2014` est considérée comme un intervalle de `24H`. Pour préciser le début et/ou la fin d'un intervalle, il faut utiliser le concept `During` avec les propriétés `hasBegin` et/ou `hasEnd`. Pour modéliser une durée, le concept `During` est aussi utilisé avec les propriétés `hasBegin` pour spécifier le début et `hasDuration` pour la durée.

Le concept `Cycle` sert à modéliser les intervalles non-convexes (répétitifs). Un intervalle non-convexe est caractérisé par deux entités: la fréquence de répétition et l'occurrence de l'intervalle convexe à répéter. Ainsi, le concept `Cycle` est relié à sa fréquence par la relation `every`. Cette fréquence est un sous-concept de `TemporalUnit` qui représente l'unité de temps à laquelle le cycle se répété. L'intervalle convexe est relié au concept `Cycle` par la relation `exp`. La propriété `sample` permet de modéliser pour les `Cycle` des échantillons de date comme `tous les 8 heures` (cf. Exemple 2b).

Notons que les propriétés `hasDate`, `hasDuration`, `hasBegin` et `hasEnd` sont des spécialisations de la propriété `hasDatation`.

Avec la modélisation proposée, nous faisons la distinction entre les intervalles infinis et les intervalles fermés. De ce fait, si les deux propriétés `hasBegin` et `hasEnd` sont spécifiées ou la propriété `hasDate` est utilisée, nous avons un intervalle fermé. Si l'une des propriétés `hasBegin` ou `hasEnd` est omise, nous avons un intervalle infini.

```
a. The first Sunday of every April.          b. Every 8H for 10 days starting from today
[a :Cycle;                                   [a :During;
   :every [a :Year];                            :hasBegin [a :Day;
   :exp  [a :During;                                       :hasDay[a: Today]];
           :hasDate                             :hasDuration [a :Duration;
             [a :Date;                                        :hasDay
                :hasDay                                         [a :Day;
                  [a :Sunday;                                    :value 10]];
                   :week 1];                   :exp [a :Cycle;
                :hasMonth                             :every [a :Hour];
                  [a :April]]]].                      :sample 8]].
```

EXMPLE 2 – *Modélisation d'intervalles non-convexes[2].*

### 3.3 HuTO et Intervalles d'Allen

Allen (Allen, 1984, 1981) définit une algèbre de 13 relations pour permettre de positionner des intervalles convexes les uns par rapport aux autres et d'en déduire des relations. Dans le sens d'Allen, un intervalle convexe est fermé et ordonné. Ainsi, il définit six paires de relations inverses: `before/after, during/contains, meet/metBy, start/startedBy, finishes/finishedBy` et `overlaps/overlappedBy`. Ainsi, à chaque fois que l'une des relations est vraie son inverse l'est aussi. La treizième relation, `equal`, est son propre inverse.

Dans HuTO, nous n'avons représenté pour l'instant que les relations `before` et `after`. Elles nous permettent de modéliser la représentation du temps en datation relative c'est-à-dire que la dimension temporelle d'une ressource est exprimée par rapport à la dimension temporelle d'une autre ressource (Exemple 3b). Notons que cette utilisation nous permet d'avoir deux informations implicites (cf. section 4.2): la date de la ressource référencée et les deux relations d'Allen entre les ressources. HuTO permet de spécifier les relations `before` et `after` entre intervalles, entre ressources et entre une ressource et un intervalle.

## 4 Annotation Temporelle des Données

L'annotation temporelle des données consiste à lier une donnée (une ressource, un triplet ou un graphe nommé) à sa dimension temporelle. Grâce à notre modélisation deux dimensions peuvent co-exister: une temporelle et une non temporelle. La dimension temporelle est spécifiée à l'aide des concepts de HuTO et la dimension non temporelle, celle du domaine de connaissance, est spécifiée au travers des triplets décrivant des aspects autres que ceux temporels.

L'annotation temporelle peut être associée à une ressource, un triplet ou un graphe nommé. Ces derniers sont temporellement annotés à l'aide de la propriété `exp` qui relie un intervalle convexe (`During`) ou non-convexe (`Cycle`) à un `TemporalThing`, elle même associée aux connaissances à annoter comme suit:

- S'il s'agit d'une ressource, le `TemporalThing` a pour valeur (`rdf:value`) la ressource concernée (cf. l'exemple 3b pour un intervalle convexe et l'exemple 3a pour un intervalle non-convexe);
- S'il s'agit d'un triplet, nous utilisons une réification RDF sur le triplet pour le `TemporalThing` (cf. exemple 3d);

---
[2]Dans l'exemple 2b, le contexte du Today a été omis volontairement pour ne pas surcharger l'exemple

- S'il s'agit d'un graphe nommé, nous utilisons `Graph`, sous concept de `TemporalThing`, dont la propriété `uri` pointe sur l'URI du graphe nommé (cf. exemple 3c.).

a. Le premier Samedi de chaque mois de Décembre, le Fanal de Ndar est organisé
```
[a :Cycle;
    :every [a :Year];
    :exp [a :During;
          :hasDate
           [a :Date;
            :hasDay
              [a :Saturday;
               :week 1];
            :hasMonth
              [a:December]];
       :exp
        [a :TemporalThing
           rdf:value
             <FanalOfNdar>]]].
```

b. la Bataille de Dekheulé a eu lieu après la Bataille de Mékhé.
```
[a :During;
   :after [a :Period;
           rdf:value
             <BattleOfMekhe>];
   :exp
    [a :TemporalThing;
       rdf:value
         <BattleOfDerkheule>]].
```

c. En 2011 la Commune de Dakar compte 1056009 d'habitants, c'est la plus peuplée et son maire est M. Sall.
```
[a :During;
    :hasDate [a :Date;
              :hasYear
                [a :Year;
                 :year 2011]];
    :exp[a :Graph;
         :uri
           <http://example.org/g/>]].

<http://example.org/g/> {
   <Dakar> <population> 1056009;
           <rang>        1;
           <mayor>       <Sall>}.
```

d. Senghor a été le Président du Sénégal de septembre 1960 à décembre 1980
```
[a :During;
    :hasBegin [a :Date;
               :hasMonth
                 [a :September];
               :hasYear
                 [a :Year;
                  :year 1960]];
    :hasEnd [a :Date;
             :hasMonth
               [a :December];
             :hasYear
               [a :Year;
                :year 1980]];
   :exp
      [rdf:subject   <Senghor>;
       rdf:predicate <presidentOf>;
       rdf:object    <Senegal>]].
```

EXMPLE 3 – *Annotation temporelle des données.*

Notons que HuTO permet aussi d'utiliser une ressource comme une référence temporelle grâce au concept `Period`. Ainsi une fois datée, une ressource peut être utilisée comme un marqueur temporel (exemple 3b).

L'utilisation de HuTO présente certains avantages comparée aux approches présentées dans la deuxième section. Dans la modélisation des expressions temporelles, HuTO permet de représenter des énoncés complexes comme dans l'exemple 2b. HuTO intègre aussi la modélisation des intervalles fermés et infinis et permet aussi d'utiliser une ressource comme référence temporelle. Ces aspects ne sont pas considérés par les autres approches présentées. HuTO permet également de modéliser le temps déictique ce que ne font pas les autres approches excepté CNTRO qui le modélise dans une chaîne de caractère.

Pour l'annotation temporelle des données, HuTO propose une représentation qui permet d'annoter une ressource, un triplet ou plusieurs triplets dans un graphe nommé. Ceci nous permet de séparer la partie temporelle des données de celles du domaine contrairement aux autres approches, excepté Temporal RDF, où la sémantique des triplets est perdue par l'introduction d'un blank node (n-ary relations, la réification RDF) ou par le déplacement des relations sur des `timeSlice` (4D-Fluents). La principale différence de HuTO avec Temporal RDF est que ce dernier nécessite une extension légère du vocabulaire de RDF (Hurtado et Vaisman, 2006).

## 5 Raisonnement Temporel et Règles

HuTO fournit un modèle conceptuel en RDFS pour modéliser des expressions temporelles et pour annoter des ressources en RDF. Cependant beaucoup de relations temporelles sont exprimées implicitement dans les occurrences d'événements. Les réponses à de nombreuses questions axées sur le temps ne sont pas nécessairement représentées explicitement mais peuvent être déduites. Pour cela, nous avons proposé un ensemble de règles permettant de normaliser la représentation des données temporelles mais également des règles d'inférences et d'implications.

### 5.1 Normalisation de la Représentation Temporelle

Puisque HuTO est une ontologie en RDFS, nous avons proposé des règles, exprimées sous forme de requêtes CONSTRUCT en SPARQL et ayant pour objectif de déduire et d'expliciter le maximum d'information temporelle afin de permettre le raisonnement sur les données.

Les informations temporelles peuvent être exprimées de différentes façons. Par exemple, une date peut être représentée soit en utilisant la représentation calendaire (Exemple 1a), soit à l'aide de chiffres (exemple 1b). Aussi, nous avons créé des règles pour normaliser ces deux types d'écritures. De ce fait, quelque soit le mode d'écriture utilisé, toutes les représentations possibles seront ajoutées dans le graphe des données. Nous avons également proposé deux règles pour déterminer les années bissextiles. Ces règles nous permettent, entre autre, de connaître le nombre de jours dans l'année ce qui est utile pour répondre à certaines requêtes. Nous avons aussi proposé une règle pour normaliser l'utilisation du concept `Period` en ajoutant la date correspondant à la période aux concepts utilisant `Period` comme date. Nous avons également normalisé les intervalles définis par leur durée en ajoutant explicitement la date de fin (`hasEnd`) de l'intervalle. L'exemple 4 montre une règle pour expliciter la date de fin d'un intervalle défini par sa durée. Notons qu'il existe sept règles pour normaliser les intervalles définis par leur durée (car la règle dépend du type de la durée qui peut être en siècle, en année, en mois, en semaine, en jour, en minute ou en seconde).

```
PREFIX dt: <http://ns.inria.fr/huto/>
CONSTRUCT  { ?x dt:hasEnd [  ?z ?t;
                             dt:hasYear [a dt:Year;
                                         dt:year ?o]}
WHERE {?x  dt:hasBegin ?y;
           dt:hasDuration/dt:hasYear/rdf:type dt:Year;
           dt:hasDuration/dt:hasYear/dt:value ?l
       ?y dt:hasYear/dt:year ?e;
          ?z ?t
       FILTER(?z != dt:hasYear)
       BIND(?e + ?l - 1 as ?o)}
```

*EXMPLE 4 – Règle de normalisation d'une durée exprimée en année.*

Dans cette règle nous récupérons toutes les propriétés liées à la date de début (`?y`) excepté l'année qu'on incrément de `?l-1` où `?l` est la durée de la ressource.

Nous avons aussi proposé une requête de vérification de la consistance entre les concepts `Cycle` et `During`. En effet, la granularité de la fréquence du concept `Cycle` doit être supérieure à celle de la date de l'occurrence de l'intervalle convexe. De même, si un `During` englobe un `Cycle` alors la granularité de la date du `During` doit être supérieure à celle de la fréquence du concept `Cycle`. Par exemple, dans l'exemple 2b nous ne devons pas interchanger les positions de `During` et `Cycle` puisque la granularité de `Today` est supérieure à celle de `Hour`.

Par manque d'espace, nous ne pouvons pas détailler toutes les règles de normalisations utilisées. Cependant il reste un travail de normalisation à faire par rapport aux intervalles nonconvexes puisque dans ces intervalles certaines informations sur les dates sont omises. L'idée serait de proposer un moyen de normaliser ces intervalles pour ajouter plus d'information sur le graphe des données.

### 5.2 Implications et Inférences

Puisque RDFS ne permet pas de modéliser certaines inférences de base comme la transitivité ou la réflexivité, nous avons créé des règles d'inférence à cet effet. Ainsi, nous avons défini par exemple des règles d'inférence pour la transitivité des propriétés `before+/after`. De même si une relation (`after` ou `before`) est exprimée entre deux événements (respectivement intervalles), il est nécessaire de propager cette relation entre les intervalles (respectivement ressources) concernés. Pour cela, nous avons proposé des règles de propagation.

Pour vérifier l'ordre d'englobement des concepts `Cycle` et `During`, nous avons défini une propriété `included` qui permet de hiérarchiser la granularité des dates. Ainsi nous avons explicité dans HuTO sept relations entre dates (`included(Year,Century,` `included(Month,Year), ...`). De fait nous avons défini deux règles de propagation: une règle pour la transitivité et une autre pour la transitivité par la subordination; c'est-à-dire si `included(d2,d1)`+ et `rdfs:subClassOf(d3,d2)` alors `included(d3,d1)`.

Notons que toutes ces règles d'inférences sont exprimées sous forme de requêtes CONSTRUCT en SPARQL interprétées comme des règles. Elles nous permettent d'ajouter plus d'information dans le graphe RDF soit en explicitant certaines informations (règles de normalisation) soit d'ajouter des informations implicites (règles de raisonnement). L'utilisation de ces règles dans des raisonneurs n'affectera que les implications de RDFS (RDFS entailment).

## 6   Requêtes en HuTO

L'ontologie HuTO (ontologie et règles) a été testée avec Corese (Corby et al., 2012), un moteur sémantique qui permet le traitement de ressources en RDF/S, SPARQL et un langage de règles adapté à RDF. Notre jeu de données compte 1014 triplets. L'exécution des implications de RDFS nous amène à 1660 triplets et celle de nos règles nous amène à 2378 triplets.

Nous distinguons deux types de requêtes: 1) celles qui déterminent les ressources relatives à une période ou relatives à une ressource temporellement annotée donnée et 2) les requêtes qui déterminent la période d'une ressource donnée.

### 6.1   Requêtes Temporelles sur les Ressources

La requête SPARQL suivante permet par exemple de déterminer la temporalité de la ressource `data:Gamou`.

```
PREFIX dt: <http://ns.inria.fr/huto/>
PREFIX data: <http://example.org/data/>
DESCRIBE ?x
WHERE{ {?x dt:exp+/(rdf:value|rdf:subject|rdf:object) data:Gamou}
      UNION
        {?x dt:exp+/dt:uri ?g
            graph ?g{ { data:Gamou ?p ?o} UNION { ?s ?p data:Gamou}}}
      FILTER NOT EXISTS {?x a dt:Period}
      FILTER NOT EXISTS {?j ?k ?x}}
```

*E*XMPLE *5 – Requête déterminant la temporalité de la ressource* `data:Gamou`.

Cette requête prend en compte les trois représentations qui peuvent être utilisées pour l'annotation temporelle d'une ressource. Ainsi quelque soit la représentation utilisée pour la ressource, cette requête nous permet de retrouver la temporalité de la ressource.

### 6.2  Requêtes sur les Éléments Temporels

Dans l'état actuel, nous pouvons par exemple déterminer les ressources récurrentes sur une période donnée. Dans l'exemple 6, la requête détermine les ressources mensuelles:

```
PREFIX dt: <http://ns.inria.fr/huto/>
DESCRIBE ?x
WHERE{ ?x a dt:Cycle;
         dt:every/rdf:type dt:Month
       FILTER NOT EXISTS {?x dt:sample ?t}}
```

*E*XMPLE *6 – Requête déterminant les données qui se produisent mensuellement.*

Par manque d'espace nous ne pouvons pas donner le détail de toutes nos requêtes types. Cependant nous pouvons déterminer:

- Les ressources répétitives par rapport à une fréquence donnée. Cette fréquence peut être un jour de la semaine, annuelle, mensuelle (Exemple 6), etc;
- Les ressources qui se produisent relativement (avant ou après) à une ressource donnée. Notons que cette requête ne concerne que les intervalles convexes puisque les propriétés `before` et `after` ne sont définies que pour ces types d'intervalles;
- Les ressources qui se produisent à une date donnée. Notons que pour ces requêtes nous avons besoin d'avoir le jour de la semaine comme argument de la requête pour avoir tous les résultats possibles. Par exemple, quelles sont les ressources qui se produisent le `Jeudi 1 Janvier 2015`;
- La date d'occurrence d'une ressource spécifique (Exemple 5).

Notons aussi que pour toutes ces requêtes types, les résultats sont des ressources annotées à l'aide d'intervalles convexes, non-convexes et ou du marqueur temporel `Period`.

Les requêtes types qui restent à traiter sont celles qui déterminent les ressources pour une période (intervalle de temps) donnée. Pour ces requêtes, la prise en compte des intervalles non-convexes est plus complexe.

# 7 Conclusions et Perspectives

Dans cet article, nous avons présenté HuTO qui est une ontologie en RDFS pour annoter temporellement des données en RDF à l'aide d'expressions du langage courant. Ce travail repose sur deux domaine de recherches dans la modélisation temporelle en Web sémantique.

Dans le domaine de la modélisation des expressions temporelles, HuTO permet la modélisation d'énoncés de temps complexe (exemple 2b). Notre ontologie comprend également un ensemble de règles afin de normaliser et de renforcer la cohérence des données temporelles. Dans notre approche nous considérons toute entité temporelle comme un intervalle pouvant être défini à l'aide de différentes granularités calendaires. Une correspondance existe entre le type `xsd:duration` et HuTO et entre le type `xsd:dateTime` et HuTO excepté la partie fuseau horaire. Une distinction est faite entre les intervalles fermés et infinis de même entre les intervalles convexes et non-convexes. Notre ontologie intègre aussi les relations temporelles `after` et `before` telles que définies dans (Allen, 1983 et 1984). Ces relations sont définies soit entre deux intervalles soit entre deux ressources soit entre un intervalle et une ressource. HuTO permet aussi d'utiliser une ressource comme un marqueur temporel pour dater une autre ressource. Une des contributions majeures de HuTO est la modélisation des intervalles non-convexes de façon à permettre l'écriture de requêtes SPARQL qui permettent de considérer tout type d'intervalle.

Pour l'annotation temporelle des données, HuTO propose une approche qui permet d'associer une dimension temporelle aux connaissances du domaine. HuTO permet également de garder la traçabilité des changements temporels sur les données que ce soit une ressource, un triplet ou un ensemble de triplets.

Plusieurs directions de recherche restent à explorer. À très court terme, nous souhaitons pouvoir traiter tous les types de requêtes concernant les intervalles (cf. section 5.2). Nous comptons traiter aussi les exceptions dans les intervalles non-convexes grâce aux graphes nommés. Nous comptons aussi étudier les autres relations définies dans (Allen, 1983 et 1984) et intégrer celles qui sont pertinentes pour le domaine socioculturel. À moyen terme, nous souhaitons proposer également des patrons de requêtes et des formats de réponses plus lisibles aux utilisateurs qui ne sont pas des experts RDF/SPARQL. À long terme, nous souhaitons enfin intégrer dans HuTO la modélisation de l'incertitude dans les notions temporelles comme dans l'expression `à la fin des années 1980`.